\documentclass{article}
\usepackage{ijcai23}
\usepackage{subfigure}

\usepackage{amsmath,amssymb,amsfonts}
\usepackage{times}
\usepackage{soul}
\usepackage{url}
\usepackage[hidelinks]{hyperref}
\usepackage{color}
\usepackage[utf8]{inputenc}
\usepackage[small]{caption}
\usepackage{graphicx}
\usepackage{amsmath}
\usepackage{amsthm}
\usepackage{booktabs}
\usepackage{algorithm}
\usepackage{algorithmic}
\usepackage{flushend}
\newtheorem{definition}{Definition}

\usepackage[switch]{lineno}

\title{Towards Interpretable Federated Learning}

\author{
Anran Li$^1$, 
Rui Liu$^2$, 
Ming Hu$^2$, 
Yuanyuan Chen$^2$, 
Shipeng Wang$^{3,4}$,  
Lizhen Cui$^{3,4}$, 
Han Yu$^{2}$\thanks{Corresponding authors.}
\affiliations
$^1$Department of Biomedical Informatics and Data Science, School of Medicine at Yale University, USA\\
$^2$School of Computer Science and Engineering, Nanyang Technological University, Singapore 639798\\
$^3$School of Software, Shandong University, Jinan, China, 250100\\
$^4$Joint SDU-NTU Centre for Artificial Intelligence Research, Shandong University, Jinan, China, 250100
\emails
anran.li@yale.edu, \{rui.liu, hu.ming, anhtuan.luu\}@ntu.edu.sg, wangshipeng95@mail.sdu.edu.cn, \\clz@sdu.edu.cn, han.yu@ntu.edu.sg
}

\begin{document}

\maketitle

\begin{abstract}
Federated learning (FL) enables multiple data owners to build machine learning models collaboratively without exposing their private local data. 
%Existing research in FL mainly focuses on improving the performance of the trained models in the face of various challenging scenarios (\emph{e.g.}, data heterogeneity). However, i
In order for FL to achieve widespread adoption, it is important to balance the need for performance, privacy-preservation and interpretability, especially in mission critical applications such as finance and healthcare. 
%there is strong demand for FL models capable of explaining the prediction results, supporting model debugging, providing insights into the contributions made by individual clients or data samples, which in turn, is crucial for allocating rewards fairly to motivate active and reliable participation in FL. 
Thus, interpretable federated learning (IFL) has become an emerging topic of research attracting significant interest from the academia and the industry alike. 
%Currently, there is no comprehensive survey on this topic. 
Its interdisciplinary nature can be challenging for new researchers to pick up. In this paper, we bridge this gap by providing (to the best of our knowledge) the first survey on IFL. We propose a unique IFL taxonomy which covers relevant works enabling FL models to explain the prediction results, support model debugging, and provide insights into the contributions made by individual data owners or data samples, which in turn, is crucial for allocating rewards fairly to motivate active and reliable participation in FL.
Specifically, we categorize methods of IFL based on the FL training process, such as client selection, sample selection, model optimization and contribution evaluation, while taking the analysis of stakeholder and privacy protection into considerations.
We conduct comprehensive analysis of the representative IFL approaches, the commonly adopted performance evaluation metrics, and promising directions towards building versatile IFL techniques.

%   The {\it IJCAI--22 Proceedings} will be printed from electronic
\end{abstract}

\section{Introduction}
Federated learning (FL) has been proposed to enable multiple data owners (a.k.a. FL clients) to collaboratively train machine learning models while preserving local data privacy \cite{mcmahan2017communication,hu2023aiotml}. Based on the distribution of local data, there are two main categories of FL scenarios: 1) horizontal federated learning (HFL), and 2) vertical federated learning (VFL). Under HFL \cite{yang2019federated,hu2023gitfl}, data owners’ local datasets have little overlap in the sample space, but large overlaps in the feature space. Under VFL \cite{yang2019federated}, data owners' local datasets have large overlaps in the sample space, but little overlap in the feature space. FL has been adopted by a wide range of applications, including financial services \cite{li2024dynamic,liu2023efficient},  smart healthcare \cite{liu2022contribution} and Industry 4.0 \cite{Chen-et-al:2023IAAI}. 
In the financial application, FL addresses several critical technical challenges: it enables collaboration across institutional data silos without exposing sensitive customer records \cite{chen2024integration}. This paradigm helps institutions jointly learn from rare but high-impact events (\emph{e.g.}, fraud or default) that are  too sparse for any single institution to capture. It also mitigates heterogeneity in distribution across institutions, improving model generalization and robustness. 
One representative application is cross-institution credit-card fraud detection, where imbalanced fraudulent cases reside at different institutions and centralized collection is infeasible; FL allows these institutions to build stronger, privacy-preserving global models for fraud and credit-risk assessment \cite{li2024dynamic,liu2023efficient}.

Today's FL models are often built on highly complex non-linear base models, \emph{e.g.}, deep neural networks (DNNs), and usually contain millions of parameters. 
The high non-linearity and complexity make it difficult for FL stakeholders to understand the internal working mechanisms of the models and decision making processes of the FL frameworks. 
This lack of interpretability may diminish trust for this emerging technology, hindering wider adoption.  
Besides, the European Parliament adopted the General Data Protection Regulation (GDPR) \cite{li2019impact} which confers a right of explanation for all individuals to obtain ``meaningful explanations of the logic involved" for automated decision making also illustrates the necessity of interpretability. 
To overcome these issues, interpretable federated learning (IFL) models capable of explaining the rationale behind model behaviors or provide insights into the working mechanisms of the models are in high demand, especially in mission critical applications such as finance and healthcare.

As a promising technology to enhance system safety and build trust among FL stakeholders, IFL has attracted significant research interest from the academic and the industry in recent years.   
Compared to the current interpretable artificial intelligence (AI) methods designed for centralized machine learning \cite{li2020survey}, IFL is more challenging due to the invisibility of local data to outsiders and the resource constraints in terms of local computation and communication power. 
It is an interdisciplinary field as it requires expertise from machine learning, optimization, cryptography and human factors in order to build viable solutions. This makes it challenging for researchers new to the field to grasp the latest development. Currently, there is no survey paper on this important and rapidly developing topic. 

To bridge this gap, we provide (to the best of our knowledge) the first survey of the IFL literature in this paper. We propose a unique IFL taxonomy which covers highly relevant works enabling FL models to explain the prediction results, support model debugging, and provide insights into the contributions made by individual data owners or data samples, which in turn, is crucial for allocating rewards fairly to motivate active and reliable participation in FL.
We put forward this survey with the following contributions.
\begin{itemize}
    \item We propose a taxonomy of IFL that comprehensively considers the FL training process, stakeholder analysis and the privacy protection analysis. 
    \item We conduct systematic analysis of the representative IFL approaches and the commonly adopted performance evaluation metrics to empirically evaluate the performance of IFL algorithms, thereby, providing readers with useful guides on experiment design. 
    \item We outline promising directions towards building versatile IFL technologies. For each direction, we analyze the limitations in the current literature and propose potential ways forward. 
\end{itemize}

%To bridge this gap, in this work, we provide a comprehensive survey of research works towards an IFL framework. We propose a unique taxonomy of IFL that organises existing works according to the FL training process, the specific interpretation techniques, the stakeholders involved and the privacy protection techniques adopted, providing a multi-perspective view into this field. We analyse the limitations of current approaches, summarise the commonly adopted performance evaluation approaches, and offer promising future directions leading towards IFL frameworks. 

% A fundamental issue in FL systems is the interpretability of FL, which has the following siginificant benefits in supporting the emergence of sustainable FL ecosystems. First, it helps to understand where the model comes from by explaining the prediction of the current global model in terms of clients and their individual samples. Second, it provides quantitative insights into the roles of individual clients in FL, and informs us whether the existence of a certain client benefits the global model. 

% It can be used to support the emergence of sustainable FL ecosystems through additional mechanisms such as reputation and incentives. 

% Without understanding and reasoning the relationships behind the predictions, these models cannot be understood and fully trusted, which prevents their applications in critical areas. This raises the need of investigating the explainability of deep graph models.

\label{sec:intro}

% \section{Preliminaries}
% \input{pre.tex}
% \label{sec:pre}

\section{The Proposed IFL Taxonomy}
In this section, we provide an overview of the proposed IFL taxonomy, discussing its structure from the perspectives of the stakeholders and the need for privacy protection.
\begin{figure*}[!t]
	\begin{center}
		\includegraphics[width=0.98\linewidth,clip]{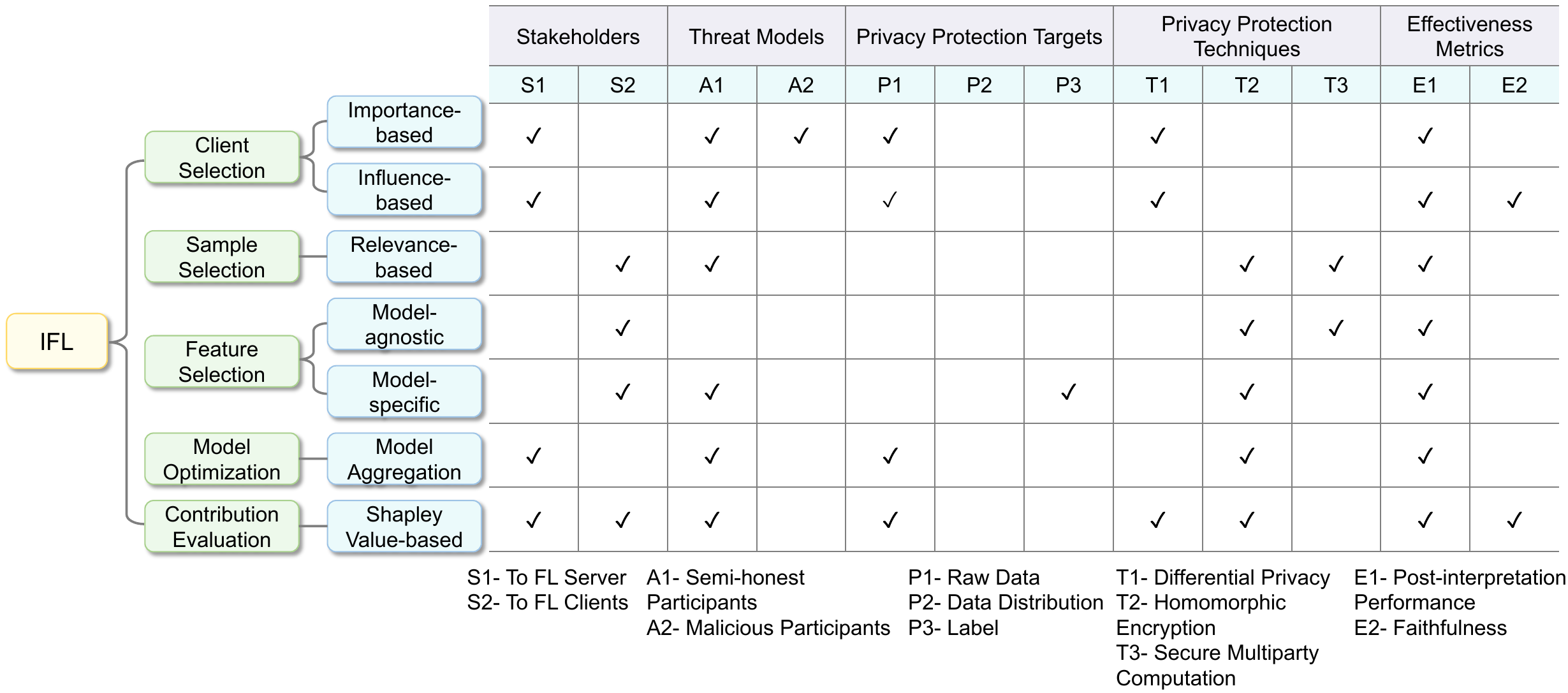}
 		\caption{The proposed taxonomy for the IFL literature.}
		%\vspace{-0.2in}
		\label{fig:taxonomy}
	\end{center}
\end{figure*}

\subsection{Definition of Interpretable Federated Learning}
Before delving into our survey, we first need to define what counts as an IFL approach. In \cite{qin2023reliable,chen2021fed}, interpretability is defined as the ability of an FL client or the FL server to select appropriate partners for cooperation or to evaluate feature contributions without revealing potentially private data, thereby leveraging interpretable learning strategies and achieving improved training results. However, this definition is strictly limited to the client selection tasks of personalized HFL settings or decision tree-based VFL settings. Therefore, in this paper, we extend the definition of interpretability to various parties' behaviors in both HFL setting and VFL setting as follows: 
\begin{definition} 
Interpretability is the ability of one party to explain or to present their behaviors related to models and data in understandable terms to other parties under an FL protocol. 
\end{definition}

\subsection{Stakeholders Analysis}
In a typical FL system, there are two types of direct stakeholders: 1) the FL server, and 2) FL clients. They are directly involved in the FL training processes. In general, under the coordination of the FL server, clients collaboratively train an FL model by sharing their local models (in a variety of forms) trained on their local datasets. Apart from them, there could be indirect stakeholders involved in IFL who may be interested in obtaining explanations regarding the FL model or the FL training process. These may include researchers and developers, regulatory agencies, policymakers and civil societies, \emph{etc}. 
Stakeholders may require different functionalities with regards to IFL. 
For instance, the FL server may need to know why a client has selected the specific data samples or features for a given FL task, and the rationale behind a specific prediction by the FL model.
The FL clients may need to know the basis on which they are being selected or excluded by the server for a given FL task, as well as the rationale behind allocating them a certain reward for their efforts. 
Researchers and developers may be interested to know under what conditions an FL model might fail so as to help with debugging. Regulatory agencies, policymakers and civil societies may be interested to know how well a given FL training framework complies with the current regulations. The diverse needs of the stakeholders require different IFL techniques.

 \subsection{Privacy Protection Analysis}
\paragraph{Privacy Protection Targets:}
In FL, the invisibility of local data is key to protecting privacy. On the other hand, it makes achieving interpretability a challenge. 
The diverse stakeholder needs also require different data privacy protection targets.  
For explaining to the FL server, the interpretable models should preserve each participant's data privacy. That is: 1) the local training data and their distributions should not be exposed to any party other than their original owners; and 2) the local data cannot be obtained or inferred by any party other than their original owners.
For explaining to the FL clients, in addition to the above two privacy protection requirements, IFL should also protect a client's labels from being exposed to or inferred by any party other than the FL server under VFL. 
%To fulfill these privacy requirements, various privacy protection techniques are needed for IFL models. 

\paragraph{Threat Models:}
Existing IFL works are often based on the following threat models. 1) \emph{Semi-honest FL participants}: they follow the FL training protocol (\emph{e.g.}, truthfully upload their local model parameters, do not collude with one another), but try to infer other clients' private information. 
% are curious to know about
%The semi-honest assumption is reasonable since participants aim to learn a high-performance FL model, and knowing other's information can be advantageous. 
2) \emph{Malicious FL participants}: the adversary can compromise FL clients and manipulate the local models during the learning process in order to compromise the global FL model to fulfil their ulterior goals. 

\paragraph{Privacy Protection Techniques:}
Following the above threat models, existing IFL works generally adopt the following privacy protection techniques: differential privacy (DP), homomorphic encryption (HE), and secure multiparty computation (MPC). 
For instance, FLDebugger \cite{li2021privacy} designs two DP-based influential sample identification methods to determine the impacts of individual training samples while preserving the privacy of clients' training data. These methods leverage the clip-based approach to bound the added noise and achieve identification performance comparable to the noise-free version. 
We summarize the privacy protection targets, threat models and privacy protection techniques adopted by IFL in the proposed taxonomy.

% Explanations might be delivered as a conversation between the explainer and explanation receivers [22]. It means that we need to consider the social context, i.e., to whom an explanation is provided [35], in order to determine the content and formats of explanations. For instance, a preferred format is verbal explanation if it is explaining to lay-users.
% \subsection{Privacy Protection Analysis}

\subsection{IFL Taxonomy Structure}
Based on the stakeholder and privacy protection analysis while considering the FL training process, we propose a taxonomy for the IFL literature as shown in Figure \ref{fig:taxonomy}. The taxonomy first identifies IFL approaches adopted by the client selection stage, sample selection stage, feature selection stage, model optimization stage and contribution evaluation stage. 
Then, it further differentiates various techniques for achieving IFL in the above stages, and highlights the stakeholders, threat models, privacy protection targets and techniques, as well as the evaluation metrics adopted by each of them to provide a concise overview of current IFL research.
%\emph{e.g.}, importance-based interpretation, influence-based interpretation and Shapley value-based interpretation, \emph{etc}. This hierarchical taxonomy provides a clear overview of the current IFL research landscape. 

% We propose a taxonomy of IFL (as shown in Figure \ref{fig:taxonomy})  according to the time when the interpretability is obtained. 

% IFL techniques can be generally be grouped into two categories: intrinsic interpretability and post-hoc interpretability. Intrinsic interpretability is achieved by constructing self-explanatory models, \emph{e.g.}, decision tree, rule-based model, linear model, attention model, which incorporate interpretability directly to their structures, while the post-hoc one requires creating a second model to provide explanations for an existing model. 
% Based on the above categorization, we further differentiate two types of interpretability: global interpretability, and local interpretability.  Global interpretability aims to understand how a global model works globally by inspecting its structures and parameters, while local interpretability examines a model's individual predictions locally, trying to figure out why the model made its decisions, \emph{e.g.}, uncover the causal relations between a specific input and its corresponding model prediction. Those two help both FL clients and FL server trust a model and trust a prediction, respectively.
% This hierarchical taxonomy provides a clear overview of the current IFL research landscape. 

\label{sec:taxonomy}

\section{IFL Approaches}
\subsection{Interpretable Client Selection}
\label{sec:client-selection}

The performance of the global FL model largely depends on the quality of the local data. 
%For example, when many participants possess erroneous data or non-IID (independent and identically distributed) data, it hinders the global model from achieving a good performance. 
IFL client selection helps the FL server understand FL model behaviours by tracing back to the distributed training datasets to identify and select clients holding high-quality data and are important to model aggregation. 
It can be achieved through: 1) importance-based techniques, and 2) influence-based techniques.

% \subsubsection{Interpretability in Relevant Client Filtering Stage}

\subsubsection{Importance-based Techniques}
These methods attempt to provide insights into the global FL model by selecting important or representative clients in each training round. There are five main approaches for calculating client importance. 

\paragraph{Model Deviation-based Methods:}
Based on the observation that local model updates from clients with noisy samples are significantly larger than normal, model deviation-based methods have been proposed to provide interpretations on the quality of local datasets. 
For FL with semi-honest participants, the deviations of the local updates from the global updates are leveraged to identify negatively influential clients with noisy samples \cite{li2021efficient}. 
For FL with malicious participants, 
\cite{NIPS2017_f4b9ec30} proposed the Krum aggregation method which selects local models similar to other local models (\emph{e.g.}, with the smallest sum of Euclidean distance) as the global model. However, Euclidean distance between two local models can be significantly influenced by a single model parameter, which can mislead Krum. To address this issue, \cite{guerraoui2018hidden} proposed Bulyan which combines Krum and a variant of the trimmed mean \cite{yin2018byzantine} to identify and aggregate high quality FL model updates received from the clients. 

\paragraph{Loss-based Methods:} Although model deviation-based methods provide decent explanations on the quality of local datasets, they cannot reflect the dynamic changes in client importance during the process of FL model training. 
Loss-based  methods employ local losses \cite{cho2020client} or loss-based utilities \cite{lai2021oort} to quantify dynamically changing client importance. The loss of a client is calculated by summing up the losses of the data samples belonging to this client. If it is larger than a threshold, the client is considered to be important and selected to participate in FL training. This approach is efficient for a client with limited resources to calculate. However, it still does not provide an accurate measure of client importance. 

\paragraph{Gradient Norm-based Methods:} 
Gradient norm-based methods provide more accurate client importance measurements compared to loss-based methods. In \cite{katharopoulos2018not}, the gradient norm of each sample of a client is computed and added up to obtain the gradient norm of the client. However, it is prohibitively expensive for a client to compute. Thus, \cite{li2021sample} leverages the gradient upper bound norm to trade off between approximation accuracy and efficiency.

\subsubsection{Influence-based Techniques} 
Influence-based methods aim to identify the impact of clients' datasets on FL model predictions. They can be divided into two categories. The first category \cite{wang2019measure,zhang2022intrinsic} perturbs or removes FL clients or their training samples to retrain the FL model. Then, the difference in performance between the new model and the original one is used to measure client influence. These methods are useful when the local datasets are similar in size and uniformly distributed. However, they become unstable in more complex FL scenarios with datasets of varied sizes and uneven distributions. Besides, since they require retraining on all clients' datasets, the evaluation process can be very expensive. 

To avoid the expensive retraining, 
influence function methods have been proposed \cite{koh2017understanding}. They use the second-order optimization technique, and generally remain accurate even as the underlying assumptions of differentiability and convexity are not holding. A straightforward method to employ 
influence functions in IFL is Fed-influence \cite{xue2021toward}. It measures the influence of a client by summing up the influence values of all its samples since the influence function has an additive property when measuring changes in test predictions \cite{koh2019accuracy}.  However, this method requires participants to directly calculate and transmit the Hessian matrix, which incurs large computation overhead (\emph{e.g.}, $O(p^3n)$ computing operations, where $p$ is the size of model parameters and $n$ is the number of total training samples) and communication overhead (\emph{e.g.}, $O(p^2 k)$ computing operations, where $k$ is the number of clients). 
To reduce the cost of influence calculation, \cite{li2021efficient,li2021privacy} leverage the Hessian vector product (HVP) to approximate the influence values, reducing the computation and communication costs to $O(np)$ and $O(pk)$, respectively.

\subsection{Interpretable Sample Selection}
In FL systems with large variety of data owned by the clients, training data may not be equally important for a given FL task \cite{katharopoulos2018not}. On one hand, it is likely that only a subset of local data from a client are relevant for the learning task, while the rest might negatively impact model training. On the other hand, among the relevant data from a client, knowledge embedded within some samples might have been extracted after some training rounds. Thus, they can be ignored afterwards without affecting final model performance. 
IFL client selection models treat all training samples of each client equally, which leads to potential waste of local computation and communication resources, and slows down model convergence. 
Therefore, IFL sample selection methods have been proposed for the server  and clients to interpret the usefulness of local data in order to improve training efficiency and model performance.
%\emph{e.g.}, identifying relevant data or filtering erroneous data. 

% prioritize samples with high relevance or greater importance to the model. Interpretability in sample selection stage aims to identify the characteristics of local data samples and provide interpretations of training samples towards predictions. 

\subsubsection{Logic-based Techniques}
Logic-based explanations connect activated concepts to illustrate the reasoning process \cite{lee2022self}. 
\cite{zhang2023lr} proposed a logical reasoning-based IFL approach to aggregate local updates with weight values determined by the quality of clients' local data. 
% They also proposed a dynamic method to determine the most appropriate logical operator for aggregating rules from multiple clients.
\cite{xing2023fedlogic} proposed a logic rule learning approach to select the optimal chain-of-thoughts prompts for improving the interpretability of federated prompt selection for multi-domain large language models (LLM). They cast this problem as a bilevel program, and solve it through variational expectation maximization. 

% \subsubsection{Relevance-based Techniques}
% Techniques have been proposed to select the subset of data samples relevant to the given FL task before starting the learning process. They usually leverage a benchmark model trained on a small benchmark dataset that is task-specific to evaluate the relevance of individual samples, and select those with high relevance. 
% Then, clients only use the selected subset of data samples in the FL model training process \cite{tuor2021overcoming}. In practice, the benchmark models are often hard to find, which limits the applicability of such techniques. 

\subsubsection{Importance-based Techniques} 
Existing importance-based IFL sample selection methods can be divided into two categories. The first evaluates sample importance based on losses. FedBalancer \cite{shin2022sample} regards samples with losses exceeding a threshold as more important samples when training the current FL model, and prioritizes them in sample selection. However, this method cannot handle erroneous samples which can also have significantly larger losses than the correct samples. 
The second leverages the gradient norm upper bound to quantify sample importance. 
For a given FL task under a budget, the server iteratively selects a subset of the most important clients, which in turn, select important local samples to build their training batches \cite{li2021sample}. To mitigate the impact of erroneous samples, a threshold (\emph{e.g.}, the median gradient norm of samples) is often adopted to filter outliers.  

\subsubsection{Influence-based Techniques}
Influence-based IFL sample selection methods have been proposed for clients to determine the impacts of individual local training samples on model predictions. Influence function methods are commonly used in determining how the model parameters change when a training point is perturbed. However, in large-scale FL systems, considering the large $n$ and $p$ in deep neural models, directly calculating influence values for all training samples will incur prohibitively high computation overhead (\emph{e.g.}, $O(np^2+p^3)$ operations) and communication overhead (\emph{e.g.}, $O(kp^2+np)$ cost). 
Thus, existing works adopt efficient influence approximation methods. One category \cite{li2021efficient,li2021privacy} leverages HVP approximation methods, which reduce computation and communication overhead to $O(np)$ and $O(pk)$, respectively. Another category \cite{rokvic2022privacy} utilizes the sign of the influence value rather than the exact influence value to measure the influence of the training samples (based on the observation that a positive influence value indicates that a data sample has a positive impact on the prediction; and vice-versa). However, this method suffers from large approximation errors when the percentage of noisy data increases. 
Recently, another work \cite{li2023fedcss} proposes a joint federated client and sample selection approach to distinguish hard samples (which are beneficial) from noisy samples (which are harmful). It is a bilevel optimization approach that performs meta-learning based online approximation to iteratively update global FL models. Theoretical analysis shows that it is guaranteed to converge in an efficient manner.

\subsection{Interpretable Feature Selection} 
The quality of clients' local features determines the effectiveness of their local models, which in turn, affect the performance of the global FL model. 
In practice, clients can possess noisy features that are irrelevant to the learning task, or a large number of redundant features which might result in model performance degradation and excessive parameter transmission. Thus, the interpretation of features (\emph{e.g.}, identifying noisy features and important features) is vital for FL. In addition, it can also provide insight into the internal working of FL models. 
% We divide IFL feature selection approaches into two categories: 1) model-agnostic techniques, and 2) model-specific techniques. 

% \subsection{Interpretability in Feature Selection Methods}

\subsubsection{Model-Agnostic Techniques}
Model-agnostic IFL feature selection approaches treat an FL model as a black-box and do not inspect the model parameters. It aims to measure the relevance of each feature to the learning task and discard the irrelevant ones. 
It can be achieved through both supervised and unsupervised interpretation methods. 

\paragraph{Supervised Interpretation:}
Supervised interpretation methods calculate per-feature relevance scores based on statistical measures (\emph{e.g.}, mutual information, Gini-impurity, F-statistics).  
\cite{cassara2022federated} proposed to iteratively identify redundant or irrelevant features in a distributed manner without exchanging any raw data. It builds on two components, a mutual information-based feature selection algorithm executed by the clients, and an aggregation function based on the Bayes theorem executed by the server. 
In \cite{li2021privacy1}, an MPC-based protocol was proposed for private feature scoring through Gini impurity, which can improve prediction accuracy while reducing model complexity.
Similarly, \cite{pansecure} proposed an MPC-based protocol for private feature relevance estimation through F-statistics to perform feature selection for VFL. 
They both follow the malicious threat model in which there is an adversary that corrupts no more than half of the participants.  

\paragraph{Unsupervised Interpretation:} Since clients' data are not always labeled, unsupervised interpretation methods for IFL feature selection have been proposed. In \cite{10017376}, a feature average relevance one-class support vector machine, Far-ocsvm, was proposed to detect outlier features. This is followed by a feature relevance hierarchical clustering step to gather representative features. By using a variable threshold algorithm, Far-ocsvm can handle the non-IID problem.

% For data features of local clients, not all features are great contributors when training the global model, some features even impair model performance. Thus,  there are works that propose to remove irrelevant features from the local and select the useful overlapping features from a federated global perspective. 

\subsubsection{Model-Specific Techniques} 
Model-specific techniques treat the FL models as white-boxes, and explicitly utilize the structure and intermediate parameters of the FL model to generates explanations. 
% Feature selection during FL training aims to select subset of relevant and important features while simultaneously learn the model, which is also called the embedded method. 
The least absolute shrinkage and selection operator (LASSO) is a well-known embedded feature selection method, with the goal of minimizing the loss while enforcing an $l_1$ constraint on the weights of the features. However, LASSO is restricted to the domain of linear functions and suffers from shrinkage of model parameters. In \cite{feng2022vertical}, the $l_2$ constraints are leveraged on feature weights in combination with an auto-encoder to select important features for deep VFL models. However, the $l_2$ constraint still suffers from shrinkage of model parameters, and requires post-training thresholds for useful features to be selected. To solve these issues,  \cite{li2023fedsdg,li2023efficient} proposed 
a federated feature selection approach, FedSDG-FS, which consists of a Gaussian stochastic dual-gate based on the $l_0$ constraints to efficiently approximate the probability of a feature being selected. 
% FedSDG-FS can achieve feature selection and model construction simultaneously. 

% while constructing the VFL global model. Data privacy is protected through Partially Homomorphic Encryption (PHE).

\subsection{Interpretable Model Optimization}
In the context of IFL, interpretability in model optimization can be achieved by designing inherently interpretable models or robust aggregation methods. The interpretable models directly incorporate interpretability into the model structures (either globally interpretable or providing interpretable individual predictions). Interpretable robust aggregation enables the FL server to understand the quality of clients' updates to perform quality-aware model aggregation. 
% In general, the more interpretable a model is the less accurate it tends to become. 
% Thus, trade-offs between predictive accuracy and interpretability should be made when constructing self-explanatory models or incorporating interpretability constraints into the models. 

\subsubsection{Constructing Inherently Interpretable Models} 
% Interpretable models can be constructed in two main ways: 1) constructing self-explanatory models (\emph{e.g.}, tree-based models and linear models, instead of complex and  opaque models); and 2) training models using data with interpretability constraints. 
% \paragraph{Constructing Self-Explanatory Models:}
Self-explanatory models, such as decision trees or random forests, can help enhance the efficiency and scalability of IFL. Under the HFL scenario, FedForest \cite{dong2022interpretable} leverages the Gradient Boosting Decision Tree (GBDT) model as the core classification algorithm, which is interpretable and efficient compared with neural networks (NNs).
In \cite{imakura2021interpretable}, an interpretable non-model sharing collaborative data analysis framework was built based on intermediate representations generated from individual local data samples. However, it assumes the availability of a shared public anchor dataset, which might not be always possible to find in practice. 
Under the VFL scenario, state-of-the-art frameworks employ anonymous features to avoid possible data breaches. However, this negatively impacts model interpretability \cite{cheng2021secureboost}. 
To address this issue in the inference process, \cite{chen2021fed} first observed that it is possible to express the prediction results of a tree as the intersection of results of sub-models of the tree held by all FL participants. Based on this observation, they proposed a method to protect data privacy while allowing the disclosure of the meaning of the features by concealing the decision paths.  

% In machine learning, decision tree ensembles such as gradient boosting decision trees (GBDT) and random forest are widely applied powerful models with high interpretability and modeling efficiency. 

% \paragraph{Adding Interpretability Constraints:} Alternatively, the interpretability of an FL model can be enhanced by incorporating interpretability constraints. In \cite{tong2022federated}, a method was proposed to enforce sparsity terms in order to reduce the number of features used for prediction, thereby improving interpretability and reducing computation cost for resource constrained clients. Other works impose semantic constraints on FL models to further improve interpretability. 
% For instance, the interpretable federated NNs \cite{roschewitz2021ifedavg} add the element-wise learned normalization layer to extract feature-wise interoperability information to enhance the detection and understanding of inter-client compatibility issues. 

\subsubsection{Interpretable Robust Aggregation Techniques}
Robust aggregation-based IFL can be divided into two categories based on the threat models they are designed to handle. For semi-honest FL participants, existing works usually adopt incentive-based methods.
In \cite{kang2019incentive}, the authors proposed an incentive mechanism that combines client reputation with contract theory to assign higher weights to high-quality updates, while motivating high-reputation clients to participate in FL.
However, it makes assumptions that the server has knowledge about clients' data quality and computation resources to enable the server to design high-paying contracts only for high-quality clients. Such information is difficult to reliably obtain in the context of FL.
To address this limitation, \cite{pandey2020crowdsourcing,zhan2020learning} proposed Stackelberg game-based incentive mechanisms, in which the server allocates rewards to the clients with the goal of achieving optimal local accuracy; while each client individually maximizes its own rewards subject to cost constraints. Nevertheless, these methods can only work when the local data samples are IID. 
% To this end, the authors in \cite{nagalapatti2021game} propose a Shapley value based federated averaging algorithm that empowers the server to select relevant clients with 

For malicious FL participants, the general approach attempts to assign lower aggregating weights to outlier clients. In \cite{li2022robust}, an aggregation method that mitigates the influence of Byzantine clients was proposed to assign lower weights to such clients. The robustness of the Byzantine client estimator in this method was also analyzed with influence values based on the observation that lower influence values correspond to stronger resistance against outliers. 

% \subsection{Interpretability in Model Optimization}

\subsection{Interpretable Contribution Evaluation}
IFL contribution evaluation can be achieved by assessing the clients' contributions to the performance of the final FL model, and assessing the contributions of the features towards a specific model prediction. 
The former is related to IFL client selection research, as the results of IFL client contribution evaluation is often used as a basis for incentivizing clients to participate in FL training and updating their reputations in preparation for future rounds of client selection. 

% In contrast, the post-hoc one requires creating a second model to provide explanations for an existing model. The main difference between these two groups lies in the trade-off between model accuracy and explanation fidelity. Inherently interpretable models could provide accurate and undistorted explanation but may sacrifice prediction performance to some extent. The post-hoc ones are limited in their approximate nature while keeping the underlying model accuracy intact.

\subsubsection{Client Contribution Evaluation Techniques}
\paragraph{Utility Game-based Interpretation:}
Utility game-based FL client contribution evaluation measures the change in coalition utility when clients join \cite{gollapudi2017profit}. 
% In utility games \cite{gollapudi2017profit}, a group of players join a team to generate social utility, and receive their payments in accordance with their contributions. 
The most common profit allocation schemes include fair value games, labor union games, and Shapley value (SV)-based games. Fair value games utilize the marginal loss of overall utility when the player leaves the union to measure a player's utility, while labor union games measure the player's utility using the marginal gain to the overall utility when the player joins the coalition. In  \cite{nishio2020estimation}, client contribution is evaluated through the gradient-based fair value scheme. However, for this type of methods, participants' contributions are influenced by the order in which they join the federation. Therefore, SV-based approaches have been more widely adopted in IFL to perform contribution evaluation that is free from the influence of the order of joining FL.

\paragraph{SV-based Interpretation:} SV-based methods provide insights into clients' datasets through evaluating their contributions to the performance of the final FL model. SV is a classic approach for quantifying individual contributions within a group. 
It assigns each participant a unique value using its contribution to the utility of all possible subsets to which it belongs. Existing SV-based methods for IFL can be divided into two categories based on the threat models.

For the semi-honest FL participants, existing works focus on improving computational efficiency, while maintaining SV estimation accuracy. This is because calculating the canonical SVs is prohibitively costly as the number of utility function evaluations required grows exponentially with respect to the number of FL participants.  
There are two main approaches for achieving efficient SV calculation: 1) accelerating within-round evaluations, and 2) reducing the number of rounds of sub-model evaluations required.  
For the first approach, gradient-based Shapley \cite{nagalapatti2021game} and local embedding-based Shapley \cite{fan2022fair,wang2023federated} have been proposed to reconstruct sub-models for different client permutations instead of re-training them from scratch. 
To further eliminate the computational costs, another approach proposes the truncated multi-round gradient-based SV evaluation by eliminating the unnecessary sub-model reconstructions \cite{wang2020principled,liu2022gtg}. 
This category of works assume that the FL server has direct access to the original FL model and a public test dataset. This might not always be valid in practice as the test data might be regarded by the FL clients as their own private assets. 
Hence, for the malicious case, the general approach studies the problem of secure SV calculation, which leverages either the ciphertext-ciphertext multiplications \cite{zheng2022secure} or blockchain-based secure FL framework \cite{ma2021transparent} that adopts secure aggregation to protect clients' privacy during the training. 

% In \cite{zheng2022secure}, an efficient two-server secure SV calculation protocol was proposed which utilizes a hybrid privacy protection scheme to avoid ciphertext-ciphertext multiplications between the test data and the models. An approximation method was also proposed to accelerate secure SV calcuation by identifying and skipping some test samples without significantly affecting the evaluation accuracy. Another work \cite{ma2021transparent} proposed a group-based SV computation protocol and a blockchain-based secure FL framework that adopts secure aggregation to protect clients' privacy during the training.

% In \cite{zheng2022secure}, an efficient two-server protocol was proposed in the SecSV approach. It utilizes a hybrid privacy protection scheme to avoid ciphertext-ciphertext multiplications between the test data and the models. An approximation method was also proposed to accelerate secure SV calcuation by identifying and skipping some test samples without significantly affecting the evaluation accuracy. 

% \paragraph{Incentivize-based aggregation with SVs:}
% \paragraph{Robust Aggregation:}

\subsubsection{Feature Contribution Evaluation Techniques}

\paragraph{SVs-based Interpretation:}
Existing SV-based IFL feature contribution evaluations methods \cite{wang2019measure,wang2019interpret} mainly leverage gradient-based SV estimation approaches to achieve high efficiency. However, they make strong assumptions that the FL server needs to know all specific IDs of the clients local features, and return the prediction part with all its features turned off. These assumptions violate the privacy of client data, and make them unsuitable for practical VFL applications. 

% \subsection{Locally Interpretable Models}
% Feature-based interpretation methods aim to identify the contributions of each feature in the input towards a specific FL model prediction.

\paragraph{Attention-based Interpretation:}
A common approach is to employ attention mechanisms that enable the server to interpret which part of inputs are utilized by the global FL model. In \cite{chen2020federated}, a hierarchical attention mechanism was proposed in which task-specific attentions are developed to evaluate personal feature correlations at the client level. A temporal attention layer is also created to evaluate cross-client temporal correlations at the FL server level. The final visualization of the attention weights can inform the clients and the server what features the global model is focused on when making individual predictions.

% Prototypes \& Criticisms - using real data instances to explain the dataset distribution; ii) Counterfactual explanations - explaining a prediction by searching or generating some instances with different feature values that change the prediction to a predefined output.

\paragraph{Activation-based Interpretation:}
Activation-based methods focus on extracting input features that highly activated neurons of a trained FL model.
Flames2Graph \cite{younisflames2graph} offers a personalized IFL solution for the multivariate time series classification problem. It extracts and visualizes the essential subsequences that highly activate network neurons in each client, and builds a temporal evolution graph that captures the temporal dependencies among these sequences.

\label{sec:FL_inter}

\section{IFL Performance Evaluation Metrics}
% \subsection{Evaluation Metrics}
To evaluate the performance of a given IFL approach, it is important to understand how useful the interpretations are and how expensive the interpretations are generated. Thus, at present, research in this field generally adopts two main categories of evaluation metrics on the effectiveness and efficiency of IFL approaches.  

\subsection{Effectiveness Metrics}
\paragraph{Post-Interpretation Performance:} One useful function of IFL is to further improve the performance of black-box models by adjusting the explanation models. Thus, the effectiveness of the explanations can be evaluated through changes in model performance (\emph{e.g.}, accuracy, errors in classification tasks) before and after the adjustments. In \cite{cho2020client,li2021sample}, the benefits of IFL client selection is reflected through lower error rates compared to the original FL models. 
Reasonable interpretations can assist researchers with model diagnostics, but whether greater performance improvement alone can be used to infer better interpretations is still in doubt \cite{li2020survey}.

\paragraph{Faithfulness:} A significant question in IFL is whether the important clients, samples and features identified are truly the relevant ones. To gauge the faithfulness of these explanations, a common method is leave-some-out retraining. It removes the identified important clients, samples or features according to their importance values from the explanations, retrain the FL models, and measure the changes in performance \cite{li2021efficient}. 
If the identified clients, samples or features are truly important, a significant degradation in performance is expected. 
In \cite{li2021efficient}, the influence scores for both noisy samples and positively important samples are calculated. The influence scores have been found to be significantly and positively correlated with the actual retraining changes in losses. 
% In this way, the faithfulness of influence functions can be demonstrated. 
Reasonable faithfulness metrics should give high scores to IFL approaches emphasizing the relevant FL entities, or reflecting the working mechanisms of the FL models. 

\subsection{Efficiency Metrics}
A defining characteristic of FL is that clients typically have limited computation and communication resources (\emph{e.g.}, sensors, mobile devices, edge devices), which makes the efficiency of IFL important.
According to clients' different resource consumption preferences, there are two efficiency metrics, computation cost and communication cost. 
%\paragraph{Computation Cost:}
The computation cost of an IFL technique can be evaluated by the amount of resources it requires. It has been assessed using computation time (usually measured through the number of elementary operations required) and memory storage requirements. 
%For instance, FLDebugger \cite{li2021efficient} achieves lower average computation operations for each local client by designing the HVP-based IFL approach. 
%\paragraph{Communication Cost:} 
The communication cost of an IFL technique can be measured by the amount of transmission it requires. This is often assessed by the number of bytes of the model parameters transmitted when training an IFL model. Nevertheless, existing research has not directly evaluated the quality of the interpretations generated by IFL approaches, as well as how they might impact the need for privacy preservation.
%In \cite{li2021privacy}, communication cost is evaluated by the sizes of the transmitted vectors for calculating the influence values. 

%To improve efficiency of interpretation methods, existing works usually adopt approximation methods for actual interpretations. For influence-based interpretability methods, to avoid expensive influence value calculations, the works \cite{li2021efficient,li2021privacy,liu2022gtg} leverage HVP approximation methods to reduce the cubic computational operations to linear operations, and reduce the quadratic communication overhead to linear overhead. For SV-based interpretable client contribution evaluation methods, gradient-based Shapley \cite{nagalapatti2021game,wang2020principled} has been proposed to reconstruct sub-models for different client permutations instead of re-training them from scratch. 
% There are often trade-offs between effectiveness and efficiency, 

% or reflect the rationale of black-box models.

% \subsection{Applications}

% \subsection{Client and Sample Selection}
% The work \cite{tuor2021overcoming} proposes a scheme that selects relevant clients' data to the given FL task before the FL starts. However, it requires an example dataset, which is hardly applicable at FL scenarios where the client data distributions are usually unknown.

% \subsection{Model Aggregation}

% \subsection{Client and Sample Contribution Evaluation}

% \subsection{Model Debugging}

\label{sec:application}

\section{Promising Future Research Directions}
Through this survey, we found that IFL research works today mostly focus on generating interpretation during individual FL training stages or for individual entities involved in different types of FL.
To make interpretability an integral part of future FL systems and support the emergence of sustainable FL ecosystems based on effective incentivization, we envision the following promising future research directions. 

\paragraph{Interpretable Model Approximation:} Existing IFL approaches leveraging self-explanatory models or adding interpretability constraints often lead to reductions in prediction accuracy.
Interpretable model extraction, also referred to as mimic learning, is a promising approach for enhancing interpretability while maintaining a high level of predictive performance. It can be used to approximate a complex FL model with an easy-to-understand model (\emph{e.g.}, decision trees, rule-based models, linear models). As long as the approximation is sufficiently close, the statistical properties of the complex model can be mimicked by the interpretable model. In this way, it could lead to IFL models with prediction performance comparable to non-interpretable FL models with much improved interpretability.

\paragraph{Hard Sample-Aware Noise-Robust IFL:} Existing IFL client and sample selection methods either ignore the existence of label noise, or simply use hand-crafted loss or gradient thresholds to filter out noisy clients/samples. However, in practice, prior knowledge of the thresholds for distinguishing noisy samples from clean samples is often not available. Besides, these methods cannot be used to distinguish positively influential clients/samples from noisy ones. This is because to mitigate noisy labels, samples with small training losses are preferred as they are more likely to be clean data. When attempting to identify positively influential samples, those with large training losses are preferred as they induce large changes in model parameters. Thus, designing IFL approaches that are hard sample-aware and noise-robust while preserving privacy is desired. 

\paragraph{Interpretability for LLMs:} Existing interpretation approaches are mainly designed for traditional FL tasks, \emph{e.g.}, image classification. 
They cannot be directly employed in federatedd LLM (FedLLM) tasks where there are potentially massive number of sequences and the base pattern of interest is a combinatorial object, \emph{e.g.}, integers, dates, URL strings, and phone numbers. These make evaluating and interpretating FedLLM more difficult and expensive.

% However, since there are potentially massive number of sequences involved in a test, FedLLM tests are both more difficult to express and evaluate, leading to tests with insufficient coverage. 

\paragraph{Interpretability under Complex Threat Models:} Most of the current IFL approaches are built on the simple threat model of semi-honest participants. This makes them vulnerable to situations in which the server or clients are malicious or colluding. This simplifying assumption needs to be relaxed to enable future IFL approaches to handle more realistic threats in practical applications. In addition, understanding how the adversaries may leverage the interpretations generated by IFL approaches to compromise the system is also important for IFL to be adopted by mission critical applications.

\paragraph{Privacy and Efficiency Trade-off:} The privacy-preserving techniques employed by existing IFL approaches incur high computation and communication costs. This make IFL unsuitable for FL systems consisting of resource-constrained devices (\emph{e.g.}, AIoT systems). Thus, research on trade-offs between privacy and efficiency is important for IFL to be adopted by such systems.  

\paragraph{Interpretability Evaluation:} IFL models are designed according to distinct principles and are implemented in various forms. This makes general interpretability evaluation challenging. Existing evaluation metrics have the following limitations. Firstly, for post-interpretation performance metrics, it remains doubtful that greater performance improvement alone directly indicate good interpretability. Secondly, the traditional leave-some-out retraining metrics are computationally expensive. Last but not least, none of the existing IFL evaluation metrics measures how much privacy might be exposed for a given level of interpretability achieved, which is crucial in the context of federated learning. Therefore, designing more appropriate and efficient interpretability evaluation metrics deserves further investigation. Such an undertaking will likely need interdisciplinary effort spanning AI and social sciences, and require standardization for adoption by the industry.

% \paragraph{Interpretable Personalized FL Models:}

% \paragraph{Causality-based Interpretable FL Models:}

% To avoid the overfitting of the decision tree, active learning is applied for training. 

\label{sec:future}

\section*{Acknowledgements}
This research/project is supported by the National Research Foundation Singapore and DSO National Laboratories under the AI Singapore Programme (AISG Award No: AISG2-RP-2020-019); and the RIE 2020 Advanced Manufacturing and Engineering (AME) Programmatic Fund (No. A20G8b0102), Singapore; the National Key R\&D Program of China No. 2021YFF0900800; and Shandong Provincial Key Research and Development Program (Major Scientific and Technological Innovation Project) (No. 2021CXGC010108).

% \newpage
%\appendix

\bibliographystyle{named}
\bibliography{ijcai23}

\end{document}